\documentclass[conference]{IEEEtran}

\usepackage{amsthm}
\newtheorem{definition}{Definition}

\usepackage{subcaption}

\usepackage{cite}

\ifCLASSINFOpdf
\usepackage[pdftex]{graphicx}
\else
\fi

\usepackage{tikz}
\usetikzlibrary{positioning, shapes, arrows.meta}
\begin{document}
%
\title{Games of Knightian Uncertainty as AGI testbeds}


\author{
\IEEEauthorblockA{Spyridon Samothrakis\\
IADS \\
University of Essex\\
Colchester CO4 3SQ, UK\\
ssamot@essex.ac.uk}
\and
\IEEEauthorblockA{Dennis J.N.J. Soemers\\
DACS\\
Maastricht University \\
PHS1, 6229 EN Maastricht, NL\\
dennis.soemers@maastrichtuniversity.nl}
\and
\IEEEauthorblockA{Damian Machlanski\\
CSEE \\
University of Essex\\
Colchester CO4 3SQ, UK\\
d.machlanski@essex.ac.uk}}



%


\maketitle

\begin{abstract}

Arguably, for the latter part of the late 20th and early 21st centuries, games have been seen as the drosophila of AI. Games are a set of exciting testbeds, whose solutions (in terms of identifying optimal players) would lead to machines that would possess some form of general intelligence, or at the very least help us gain insights toward building intelligent machines. 
Following impressive successes in traditional board games like Go, Chess, and Poker, but also video games like the Atari 2600 collection, it is clear that this is not the case. Games have been attacked successfully, but we are nowhere near AGI developments (or, as harsher critics might say, useful AI developments!). 
In this short vision paper, we argue that for game research to become again relevant to the AGI pathway, we need to be able to address \textit{Knightian uncertainty} in the context of games, i.e. agents need to be able to adapt to rapid changes in game rules on the fly with no warning, no previous data, and no model access.   
%

\end{abstract}


%
\IEEEpeerreviewmaketitle

\section{Introduction}
Artificial Intelligence (AI) research papers have traditionally justified their use of games as benchmarks with the claim that they provide scaffolding for Artificial General Intelligence (AGI); the paper (and blog post) by Togelius~\cite{togelius2017ai} best captures early optimism in the process. Games are cultural artefacts of considerable importance, but also excellent ``mini-universes'' one can experiment in. Although it was never clear what this AGI would look like at the limit (although certain authors have tried to be more specific~\cite{morris2023levels}, with practical levels of automation being discussed even further back~\cite{degradation}), one can assume that the goal was to end all the need for labour or to ``summon'' an entity that would solve almost all human problems. Within the wider scope of understanding intelligence, games would also play a significant role in telling us more about ourselves, thus potentially helping with advances in fields like psychology and the broader cognitive sciences. However, we have lately experienced a shift; games are no longer seen as the ``way to AGI.'' The advent of Large Language Models (LLMs) moved the interest of the AI community away from games, and more towards methods that learn from vast quantities of data through a self-supervised process.

The problem of games (and related research, such as game competitions) not being up to the task of helping bring about AGI was spotted early on by Chollet~\cite{chollet2019measure}. The insights identified never seemed to catch on with the game AI community, which was getting involved deeper into games for the sake of games (i.e. studying the artefact itself), rather than using games with the goal of AGI in mind. This paper works on the premise that, while a large part of the games community is interested in games for the sake of games, the almost universal shift to games as cultural artefacts is just a deviation stemming from perceived impotence. Drawing inspiration from recent discussions in the field of economics\cite{sunstein2023knightian}, we propose a set of example games and benchmarks that can help rebuild trust in the process of ``game AI for AGI'' agenda. In particular, we propose working around a new class of games, which we term ``Games of Knightian Uncertainty,'' where rapid rule changes are the norm. 

The rest of this paper is organised as follows: Section~\ref{sec:failure} outlines the causes behind the failure of games as AGI testbeds, Section~\ref{failures-ggp} discusses specifically why General Game Playing is no longer a useful AGI benchmark, while Section~\ref{sec:night} introduces Knightian Games. We conclude with a very short discussion in Section~\ref{sec:conc}.

\section{What problems are still hard for AI and why?}\label{sec:failure}
\begin{figure*}[htbp]

    \centering
    \begin{subfigure}[t]{0.3\linewidth}
        \centering
        \includegraphics[width=\linewidth, height=5cm]{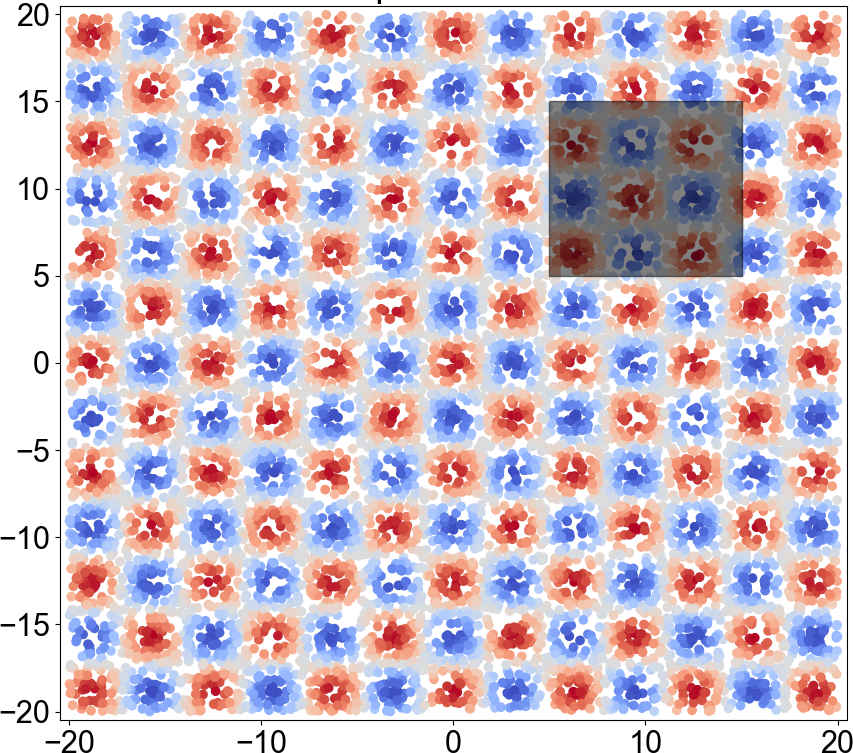}
        \caption{A board simulated using $y 
= cos(x_0)cos(x_1)$, with the color intensity denoting $y$ and $x_0$ and $x_1$ $\mathcal{U} \sim
[-20,20] \setminus [5,15]$ for both training set and test set data points, while OOD test set is $\mathcal{U} \sim [5,15]$ (the shaded area). }
        \label{fig:board0}
    \end{subfigure}
    \hfill
    \begin{subfigure}[t]{0.3\linewidth}
        \centering
        \includegraphics[width=\linewidth, height=5cm]{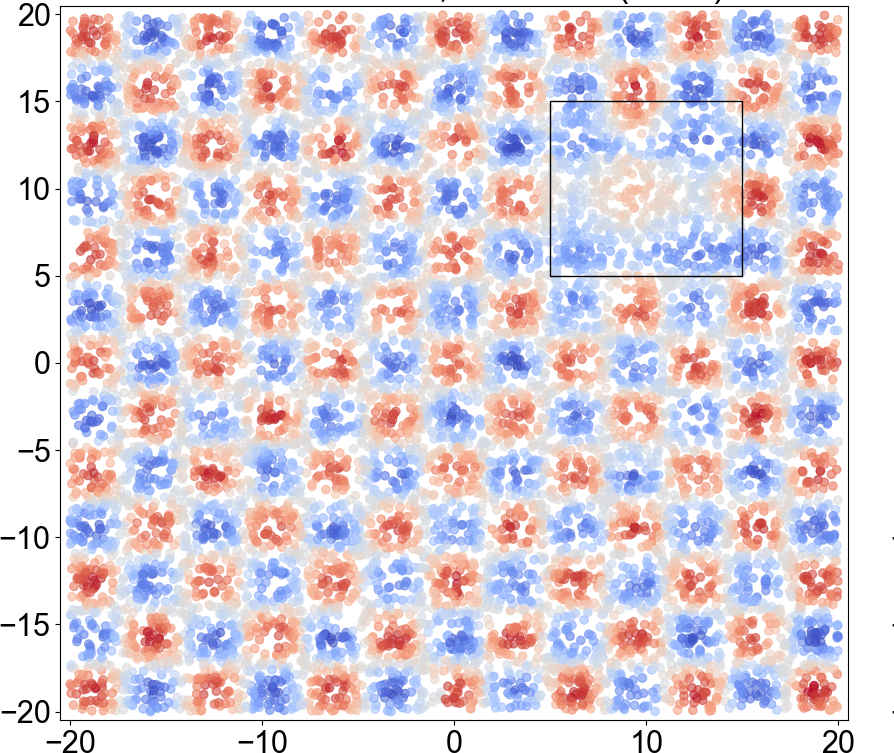}
        \caption{Results using LightGBM; test MSE outside the OOD area is $\approx 0.01$, while inside the OOD area is $ \approx 0.42$.}
        \label{fig:board1}
    \end{subfigure}
    \hfill
    \begin{subfigure}[t]{0.3\linewidth}
        \centering
        \includegraphics[width=\linewidth, height=5cm]{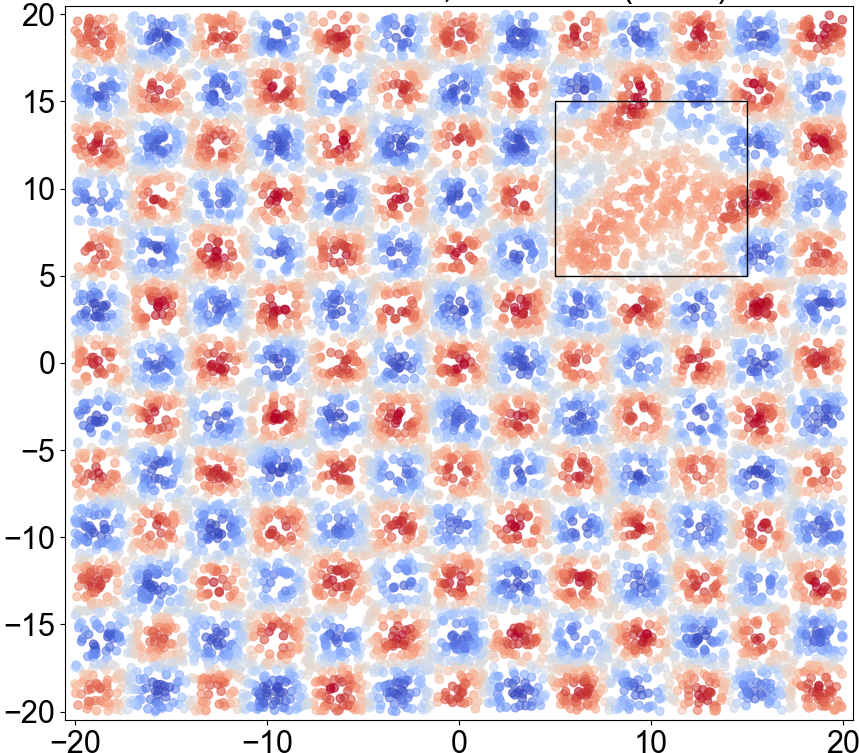}
        \caption{Results using a standard MLP (trained using AdamW, 4 layers of 20 ReLU neurons); test MSE outside the OOD area is $\approx 0.01$, while inside the OOD area is $\approx 0.29$.}
        \label{fig:board2}
    \end{subfigure}
   
    \label{fig:board}

 \caption{Training, test and OOD test data and the default responses of modern regressors. Notice the significant error on OOD test set (within the shaded area) -- the gap between data points is far too wide for modern ML methods to generalise, while a human would arguably fill in the pattern trivially. }
\end{figure*}

The recent successes of LLMs have created a situation where it is hard to argue that there is an alternative path to AGI that does not involve huge amounts of data and high GPU cycles. Yet these models fail in a multitude of tasks, which range from addition to reasoning. The status quo is so pro-LLMs that any proof of their shortcomings lies with the accuser, as the default perspective is that LLMs are indeed a form of AGI (or an AGI to be, pending some minor problems here and there). A more honest evaluation would show that the original criticisms of Chollet~\cite{chollet2019measure} and Mitchell~\cite{mitchell2021abstraction} still hold. More specifically, the problems we face come from at least two major aspects related to foundational practices in modern Machine Learning (ML), discussed in the following subsections.

\subsection{Representation learning learns incomplete representations}
The basic selling point of representational learning (i.e. what neural networks aim to do) is to help create abstractions that can be reused widely beyond the specific case. This was to be achieved through standard machine learning (given enough data) or by combining various data streams into one large network~\cite{rakelly2021mutual}. Intuitively, the discrete analogue of this research programme would be genetic programming/symbolic regression, where the basic functions are discovered by an algorithm. Unfortunately, this is almost never the case. The latent variables learnt by neural networks often do not capture any global properties meaningfully and thus fail to extrapolate, something covered extensively by Chollet~\cite{chollet2019measure} and termed ``memorising infinity'' by Saba~\cite{mediumMemorizingUnderstanding}. Representational learning, instead of creating useful abstractions, seems to be trying to fill a huge hashmap-like structure with as much knowledge as possible. An example of this is shown in Figure 1, where, while out-of-the-box regressors successfully generalise in a test set that comes from the same distribution, they fail when it comes from a different distribution. Note that symbolic regression would be successful in this case, but as we increase input dimensions, it becomes computationally impossible. The situation is bad enough that, in order to reach superhuman levels, agents have to play games orders of magnitude more than humans to fill those differentiable infinitely-large hashmaps. As a consequence of memorising instead of learning robust abstractions, quite often, and combined with other issues such as catastrophic forgetting, agents tend to lose to adversarial examples~\cite{wang2023adversarial}.

\subsection{The real world is non-stationary and open}
Openness tends to refer to different things in Reinforcement Learning (RL) and economics; in RL, the environment is considered to be somewhat stationary~\cite{stanley2017open, samvelyan2022maestro}. One is not expected to learn GO and then generalise to backgammon but at best slide into an adjacent game. That is, most open-ended AI frameworks assume that there is a series of fixed environments that do not differ dramatically from one another, and open-endedness tends to focus on agents having increasingly more complicated behaviour. In economics, openness comes from new events that appear unexpectedly, and no statistical relationship with what was happening before can be established. Imposing closed-model thinking to open systems has been argued to be catastrophic~\cite{loasby2003closed}. Static environments, where an agent tries to identify how to operate in a single reward regime~\cite{friston2007free,hutter2005universal} are the norm among AI theories, with the upward generalisation limit often set to understanding the causal structure of the world~\cite{Richens2024RobustAL}. This implies that a large portion of all possible ``realities'' would have been traversed by an agent throughout their lifetime, and at some point an agent will select how to act based on past observations. There is an argument to be made, however, stemming from failure to make sufficient progress in robotics and self-driving cars, that the world is not stationary, but it changes all the time in unexpected ways, which makes AI with statistical reasoning unsuitable as its core mechanism of analysis due to way too many ``unknown unknowns''.


\section{General Game Playing}~\label{failures-ggp}
The most obvious example of using games as ``AGI scaffolding'' comes from general game playing competitions. This includes GVG-AI~\cite{GVGAI} and GGP~\cite{finnsson2010learning}, but also allied approaches~\cite{stephenson2019overview} and Atari games. The idea here is that one has to get insights beyond certain hand-coded heuristics of playing a single game (something popular in single-game tasks) and come up with algorithms that could attack any game from scratch. The problem with these competitions is that the agents involved are not general in any sense. Almost always, the agents assume the existence of an easy-to-access (either online or offline) model, which does not adequately capture settings where more general intelligence would be needed. At best, what is learnt is how to act in a wide array of similar environments, i.e. the outcome agents are \emph{adapted} to a specific game, not \emph{adaptive} to wider environments. Overall, the independent and sequential adaptation regime that this type of research promotes might be important for algorithmic development but does very little to support AGI goals. Quoting Hernández ~\cite{hernandez2017evaluation} ``AI is the science and engineering of making machines do tasks they have never seen and have not been prepared for beforehand'', and if we assume the agent has received a perfect model a-priori, a massive chunk of the problem is effectively solved by the model. Researchers have indeed tried to soften these requirements, but with some notable exceptions~\cite{chollet2019measure}, open settings (as for example in Wang et al.~\cite{wang2020enhanced}) remain limited to very basic distributional shifts between what one uses during training and during testing. 

\section{Knightian uncertainty}~\label{sec:night}

\subsection{Types of uncertainty}
Researchers developing game playing agents have (mostly) so far focused on two forms of uncertainty; \emph{epistemic} and \emph{alleatoric}~\cite{lockwood2022review}. Both forms of uncertainty are very well known to game designers and game players. Assuming an extensive-form game (the most general form of such games), epistemic uncertainty refers to the lack of knowledge that an agent playing the game has about the world. In practical terms, this translates into a quantification of ``how good'' an action is. Given enough playthroughs, a well-adapted agent should drop this to zero; the effects of every action should be known to them. In contrast, alleatoric uncertainty reflects parts of the rules of the game where randomness is irreducible, e.g. dice rolls, drawing up cards (or, in reinforcement learning lingo, stochastic transitions, rewards and actions). In traditional games, depending on game type, there are incentives for agents to play around with both forms of uncertainty (e.g. increasing alleatoric uncertainty through bluffing in poker so as to increase the epistemic uncertainty of your opponents, or decreasing epistemic uncertainty while increasing alleatoric uncertainty through exploratory actions). Both forms have been studied extensively in RL, mainly through the exploration literature (e.g. see~Jiang et al.~\cite{jiang2023general} or Turner et al.~\cite{turner2021optimal} for excellent reviews), and attacking them is vital for all agents, but it does not address the main problems of agents learning useful abstractions and being able to act with limited data. 

\subsection{Knightian uncertainty}
Historically, a third form of uncertainty, termed Knightian uncertainty, has been both important and overlooked. The concept is borrowed from economics (for an excellent overview, see Sunstein~\cite{sunstein2023knightian}) where, as Keynes puts it, there might be situations where \textbf{ ``there is no scientific basis on which to form any calculable probability whatever''}, and it is tightly coupled with ``black swans'' or ``unknown unknowns''. The real world is an excellent source of such events, however it was never clear how these concepts could be applied to games. What we propose here is to allow out-of-distribution (OOD) events to take place habitually throughout a game.  As an example, imagine an agent playing chess but suddenly there is a change in the rules of the way the rook moves, and now it moves the same way as a bishop or a king, while the board's starting configuration is random. Whatever policies and value functions were learnt until this point, they are now moot. If one treats everything as alleatoric or epistemic uncertainty, the intrusion of new rules mid-game would cause havoc to an already trained agent, and would require (at best) extensive retraining to bring the epistemic uncertainty levels down to something manageable. To push things even further, imagine an agent taught how to play backgammon, but now forced to play checkers with the remaining pieces without having seen checkers before. Although a formal definition of what is a game of Knightian uncertainty currently does not exist, we propose the following:

\begin{definition}
  A game of Knightian uncertainty is one where the transition function, rewards and the set of actions and observations are themselves non-stationary functions that can change abruptly at any point. 
\end{definition}

\begin{figure*}
    \centering
    \begin{tikzpicture}[node distance=2cm, auto, font=\sffamily, ]
        \tikzstyle{block} = [rectangle, draw, text width=9em, text centered, minimum height=7em]
        \tikzstyle{line} = [draw, -Latex]
        
        \node [block] (original) {Training\\(e.g. a limited collection of GVG-AI games)};
        \node [block, right=3cm of original] (nearOOD) {Testing Near OOD\\(i.e. variations of the training games )};
        \node [block, right=3cm of nearOOD] (farOOD) {Testing Far OOD\\(e.g. different games to the training set)};

 
            \path [black,line,->] (nearOOD.east) to[out=60,in=120] node[above] {retrain} (nearOOD.west);
        
  \draw [line] (original) -- node[above]{5-10 demo runs} (nearOOD);
        \draw [line] (nearOOD) -- node[above]{5-10 demo runs} (farOOD);
    \end{tikzpicture}
    \caption{A potential benchmarking setup for Knightian uncertainty. Agents train using the original game and are expected to generalise to variations, while after training with variations (``Near OOD'') they are asked to generalise to different games.}
    \label{fig:game_variations}
\end{figure*}
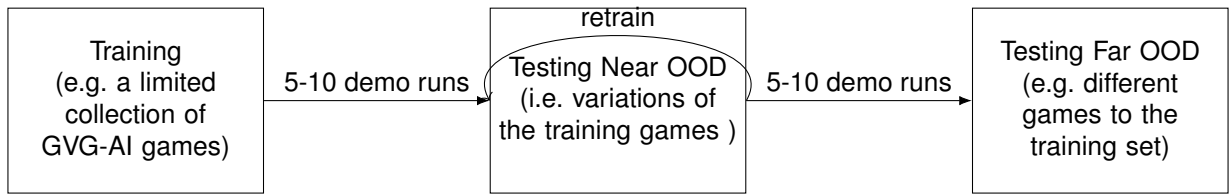

In other words, the agent is never sure not only which game they are playing (this is generally called ``incompleteness'' in game-theoretic lingo), but also that they do not know which games are potentially possible. 
One might claim that an agent might need some hints to link back to prior knowledge, but this should not come from linking back to probabilistic feature spaces (and the metric functions this alludes to) but by creating abstractions and reasoning, something which is probably embedded within human value functions. Language and the way we communicate through stories and the generalisation capacity inherent in successful models~\cite{hupkes2020compositionality}, potentially combined with causality~\cite{scholkopf2022causality},  could potentially be a way forward. To help support research on such games, a set of benchmarks should be organised to create measurable objectives, building upon GGP. One such example, portrayed in Figure \ref{fig:game_variations}, would be as follows. There are two different ``steps'' involved, in which participants are given a set of games (e.g. variations of chess), and they are asked to generalise to either different variations of the same game (e.g. different chess variants) or to completely different games (e.g. backgammon). We call the first step ``near OOD'' (and reflect games that would be reachable through domain randomisation) and the second step ``far OOD'', which includes different games -- \textbf{no model is to be provided, thus making search impossible}. During evaluation, agents would be exposed to a limited set of demo games (e.g. 5-10 demo runs) for their new setup, but otherwise no more information should be provided. The interface should follow the well-known observation-action paradigm of AI-GYM, but with both observations and actions of arbitrary size.  Example setups for a number of demo cases are provided below. 


\subsubsection{Chess}

\noindent\textbf{Training Set:} Players learn the fundamentals of chess through the training set, which represents the standard version of the game played on an 8$\times$8 grid with traditional pieces and rules.
\noindent\textbf{Near OOD:} Test set variations include Chess 960 (Fischer random chess), where the starting position of the pieces is randomised, and chess variants with alternative board sizes or pieces or new types of pieces. 
\noindent\textbf{Far OOD/Knightian:} Playing poker and backgammon after training in near OOD chess variants.

\subsubsection{Poker}
\noindent\textbf{Training Set:} Games both limit and no-limit Texas Hold'em.
\noindent\textbf{Near OOD:} Poker variants such as as Omaha, and 5-Card Draw.
\noindent\textbf{Far OOD/Knightian:} Playing Chess or Go after training in near OOD poker variants.

\subsubsection{Mario}
\noindent\textbf{Training Set:} Players learn the basic mechanics and controls of Mario Bros, a classic platformer.
\noindent\textbf{Near OOD:} Variations include different levels, challenges, and power-ups, as well as fan-made levels and mods created by the community.
\noindent\textbf{Far OOD:} Playing Pac-Man following extensive training in the near OOD setting. 

\subsubsection{GVG-AI}
\noindent\textbf{Training Set:} A small set of GVG-AI (General Video Game AI) games.
\noindent\textbf{Near OOD:} Variations of the original training set (e.g. adding new enemies to space invaders if they were included in the training set).
\noindent\textbf{Far OOD:} A different set of GVG-AI games, that have as little relationship to the near OOD setting as possible.

\section{Conclusion}\label{sec:conc}
The games community would either have to accept its irrelevance in the AGI race, or refocus to game benchmarks that matter. In tandem with other AI subcommunities, there is a tendency to ignore the harder problems precisely because they are hard, and solving them requires new approaches that are not available at the time. However, if one is to accept that games do have a role to play in setting up meaningful benchmarks, current competition setups would have to change widely and test the limits of the generalisation capacity of game-playing agents.




\bibliographystyle{IEEEtran}
\bibliography{refs.bib}

\end{document}